\documentclass{article}
\usepackage{spconf,amsmath,graphicx}

\usepackage{subcaption}
\usepackage{xcolor}
\usepackage{units}
\usepackage{multirow, makecell}
\usepackage{adjustbox}
\usepackage[flushleft]{threeparttable}

\usepackage{booktabs}
 
\usepackage[utf8]{inputenc}
\usepackage{kotex}

\newcommand{\asr}{Wav2Vec 2.0 based Seq2Seq Model}
\newcommand{\nlu}{RoBERTa}

\newcommand{\interface}{CTI}

\usepackage{lipsum}

\newcommand\blfootnote[1]{%
  \begingroup
  \renewcommand\thefootnote{}\footnote{#1}%
  \addtocounter{footnote}{-1}%
  \endgroup
}

\title{Integration of Pre-trained Networks with Continuous Token Interface for End-to-End Spoken Language Understanding}

\name{Seunghyun Seo$^{1,2}$$^{\ast}$, Donghyun Kwak$^{1*}$, Bowon Lee$^2$}

\address{$^1$Clova AI, NAVER Corp.  $^2$Inha University \\
    \small{\texttt{\{real.seunghyun.seo, donghyun.kwak\}@navercorp.com, bowon.lee@inha.ac.kr}}}

\begin{document}
%
\maketitle
%

%

\begin{abstract}
Most End-to-End (E2E) Spoken Language Understanding (SLU) networks leverage the pre-trained Automatic Speech Recognition (ASR) networks but still lack the capability to understand the semantics of utterances, crucial for the SLU task.
To solve this, recently proposed studies use pre-trained Natural Language Understanding (NLU) networks.
However, it is not trivial to fully utilize both pre-trained networks; many solutions were proposed, such as Knowledge Distillation (KD), cross-modal shared embedding and network integration with Interface.
We propose a simple and robust integration method for the E2E SLU network with a novel Interface, Continuous Token Interface (CTI).
CTI is a junctional representation of the ASR and NLU networks when both networks are pre-trained with the same vocabulary.
Thus, we can train our SLU network in an E2E manner without additional modules, such as Gumbel-Softmax.
We evaluate our model using SLURP, a challenging SLU dataset and achieve state-of-the-art scores on intent classification and slot filling tasks.
We also verify that the NLU network, pre-trained with Masked Language Model (MLM), can utilize a noisy textual representation of CTI.
Moreover, we train our model with extra data, SLURP-Synth, and get better results.

\end{abstract}


\blfootnote{$^\ast$ Equal contribution.}


\begin{figure*}[ht!]
\centering
    \subfloat[Knowledge Distillation]{%
        \centering
        \includegraphics[height=0.14\textheight]{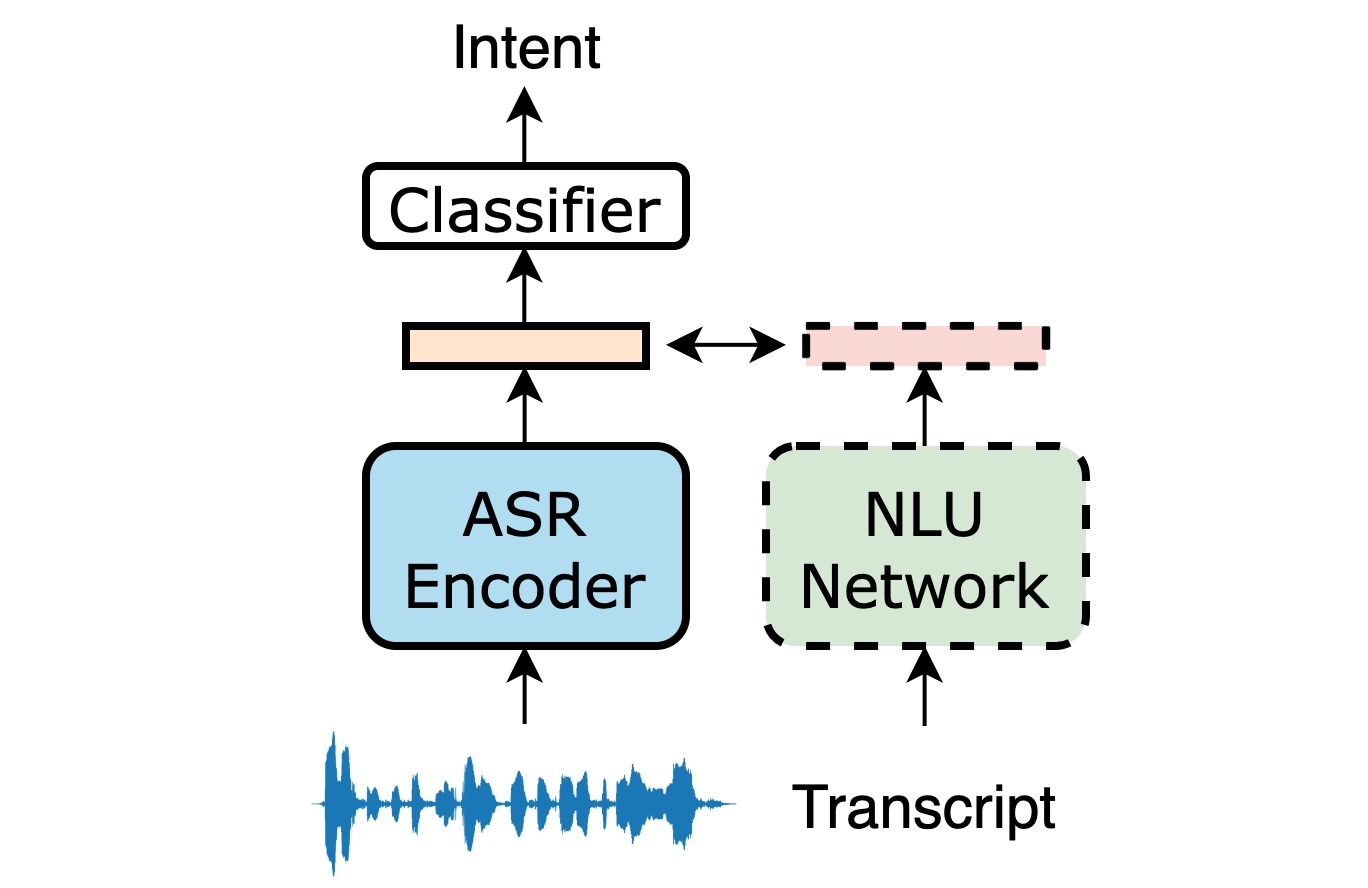}}
    \hspace{-6mm}
    \subfloat[Cross-Modal Shared Embedding]{%
        \centering
        \includegraphics[height=0.14\textheight]{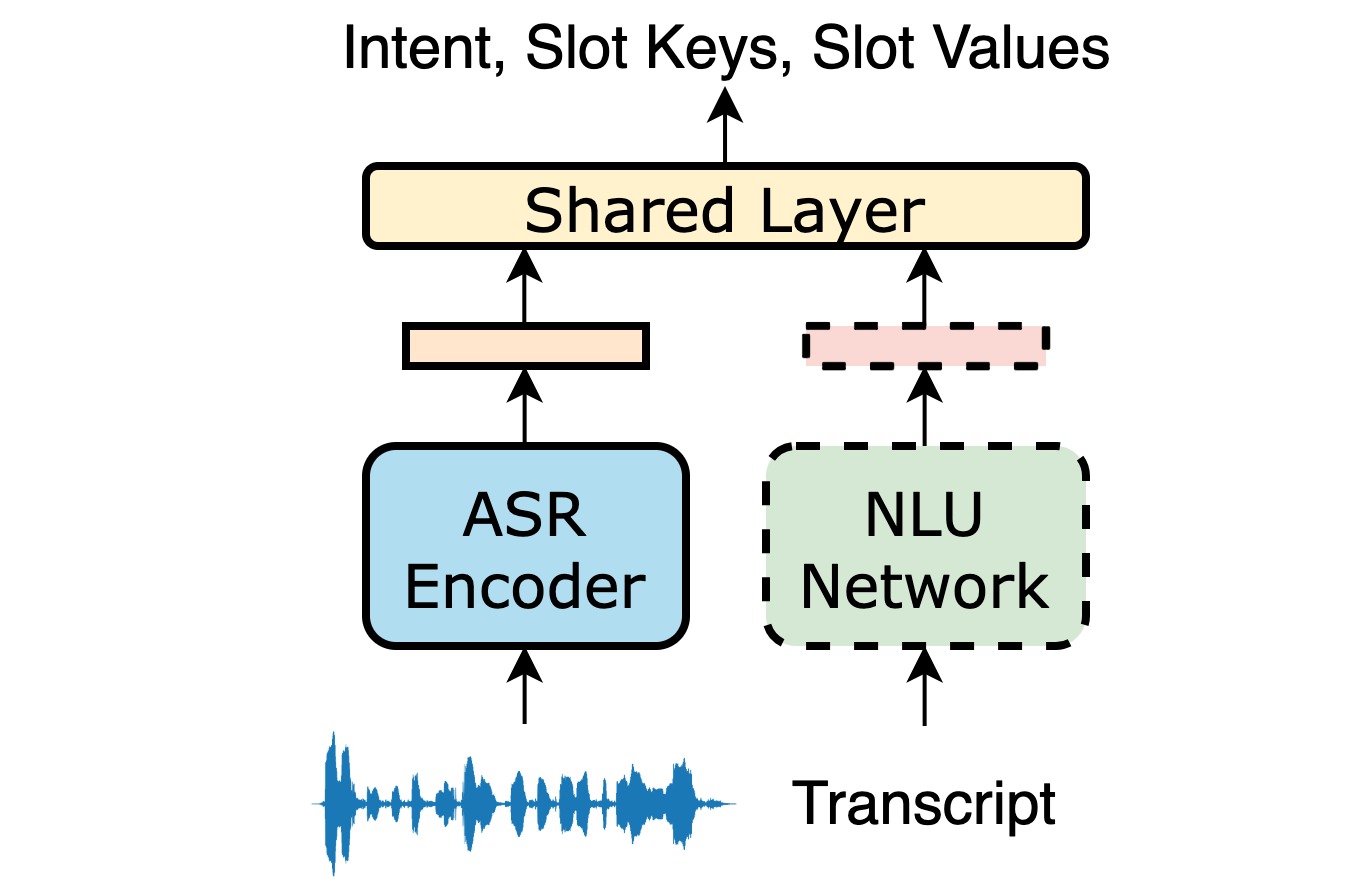}}
    \subfloat[Network Integration with Interface]{%
        \centering
        \includegraphics[height=0.14\textheight]{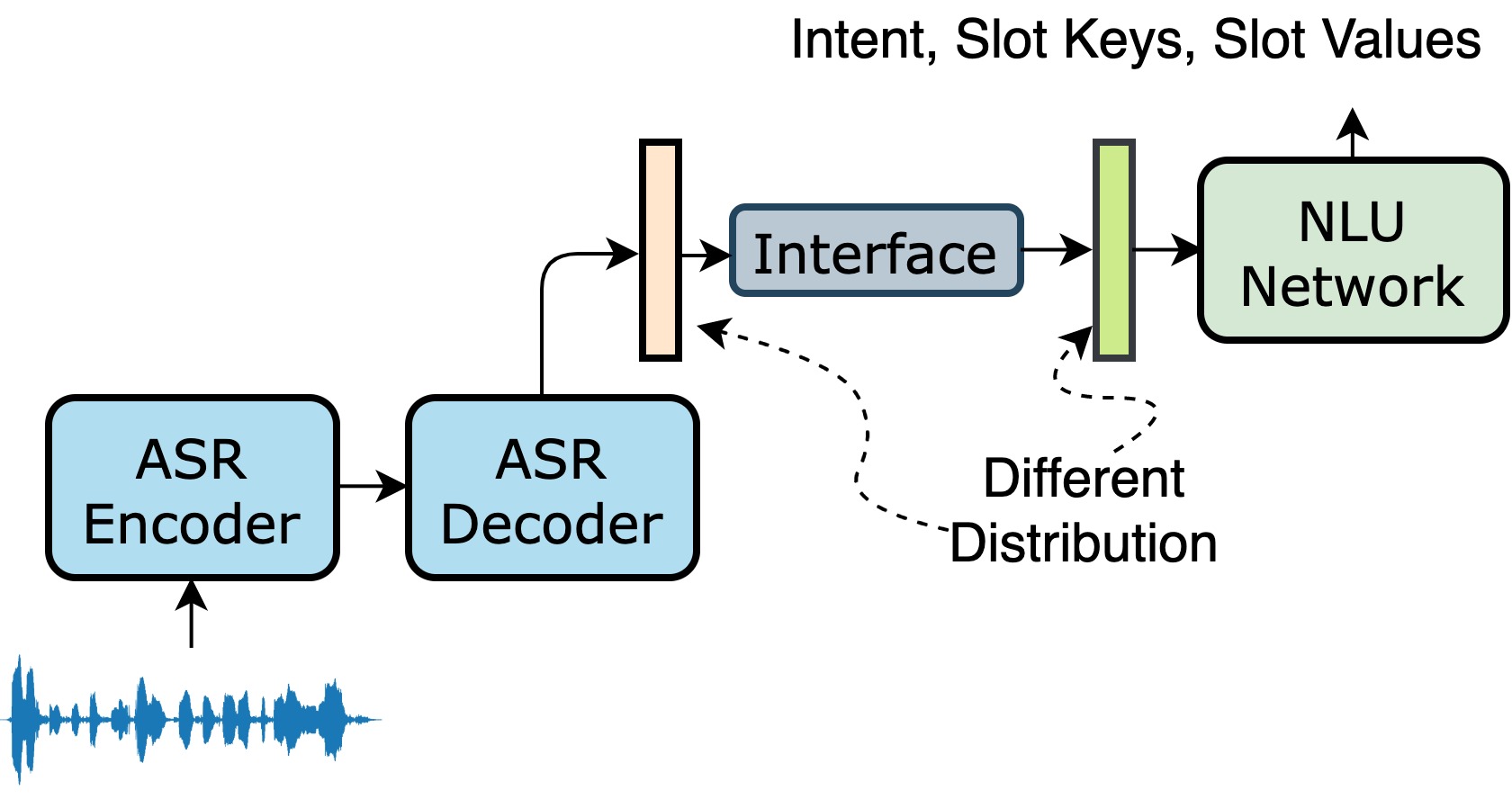}}
\vspace{-2mm}
\caption{Various methods for leveraging the capacity of the pre-trained NLU network. From left to right, they are methodologies in the order described in Related Work. 
Note that the blocks with dotted lines are only used in the training procedure.
}
\label{fig:e2e_slu_with_lm} 
\vspace{-6mm}
\end{figure*}


\vspace{-6mm}
\section{Introduction}

Spoken Language Understanding (SLU), a front-end of many spoken dialogue systems, is a task of extracting semantic information such as intents, slots, or emotions from speech. Conventional SLU pipelines consist of the Automatic Speech Recognition (ASR) model and the Natural Language Understanding (NLU) model, where the ASR model converts speech to text, and the NLU model predicts task-specific information from the text~\cite{tur2011spoken}.
However, these pipelines propagate errors from the ASR model to the NLU model, degrading the system's overall performance.
To address this problem, recent studies focused on End-to-End (E2E) SLU networks that directly extract semantic information from speech~\cite{serdyuk2018towards,Lugosch2019}. Most E2E SLU networks leverage the encoder part of pre-trained ASR networks but cannot fully understand the semantics of utterances, which is crucial for the SLU task.

To solve this, recently proposed studies use pre-trained NLU networks to leverage the powerful language representation of the Pre-trained Language Model (PLM), such as BERT~\cite{devlin2019bert}. Some approaches use Knowledge Distillation (KD) from PLM to SLU network~\cite{Cho2020,kim2021two} or introduce cross-modal shared embedding space between acoustic and textual features~\cite{agrawal2020tie,huang2020leveraging,kim2021st}, and others try to integrate the ASR and NLU networks with appropriate Interface~\cite{lai2020towards,rao2021mean}. 

However, these methods may lose the powerful language representation of BERT in the process of KD or lose important features while embedding cross-modal inputs in a shared space.
In the case of network integration with Interface, it depends on the gap between the two representations.
The Interface is a junctional representation of two pre-trained networks to be combined~\cite{rao2021mean}. 
To make the gradient flow via Interface, it is usually necessary to employ additional modules to match the representations.

For example, If we integrate the networks with Discrete Token Interface (DTI), one-hot encoding of ASR hypothesis, the model needs additional modules, such as the Gumbel-Max trick and Gumbel-Softmax distribution for forward and backward passes to match the distribution~\cite{jang2017categorical}. 
This learning process is complicated, as the pre-trained networks could miss their acoustic or linguistic representations.

In this paper, we propose a novel Interface, Continuous Token Interface (\interface), for a simple and robust network integration.
The CTI is naturally derived by pre-training both ASR and NLU networks with the same vocabulary.
With our interface, there is a small representation gap between the networks, and no additional modules are needed.
Because the networks have the same vocabulary, the softmax probability distribution of the ASR network can be considered as a noisy textual representation from the point of view of the NLU network.
We assume it only differs in the noise pattern; our Interface has a neglectable gap of representations, unlike other Interfaces.
Therefore we directly feed the ASR network's output to the NLU network, which makes our SLU network trainable in an E2E manner without additional modules, such as Gumbel-Softmax.

Finally, we evaluate our model on the SLURP dataset~\cite{bastianelli2020slurp}, a recently proposed challenging SLU dataset, and achieve state-of-the-art performance in both Intent Classification (IC) and Slot Filling (SF). 
We also conduct ablation studies to verify that the NLU network, pre-trained with Masked Language Modeling (MLM), can utilize a noisy textual representation of \interface.
Moreover, we train our model with extra data, SLURP-Synth, and get better results.




Our main contributions can be summarized as follows:
\vspace{-2mm}
\begin{itemize}
    \item[1.]{We integrate the pre-trained networks by a novel Interface, CTI, in an E2E manner without any modules.}
    \vspace{-3mm}
    \item[2.]{Our CTI integration allows each part of the model to be trained independently based on data type.}
    \vspace{-3mm}
    \item[3.]{We evaluate our model on the SLURP dataset and achieve state-of-the-art scores in both IC and SF.}
    
\end{itemize}

\vspace{-6mm}
\section{Related Works}

Recently, the E2E SLU approaches are designed to understand contextual semantic information of speech by utilizing PLM~\cite{lai2020semi,chung2021splat}.
These works were studied in various ways, including KD, learning cross-modal latent space, and network integration via Interface.
Fig. \ref{fig:e2e_slu_with_lm} shows these works.

\vspace{-4mm}
\subsection{Knowledge Distillation}
Inspired by~\cite{hinton2015distilling}, KD-based E2E SLU models were proposed to leverage the representation power of PLM~\cite{Cho2020,kim2021two}. The SLU network can learn high-level semantic representations from the NLU network by distilling the knowledge.

\vspace{-4mm}
\subsection{Cross-Modal Shared Embedding}
These approaches were designed to learn a cross-modal latent space by inducing the paired speech and transcript to become closer to each other in the common embedding space~\cite{agrawal2020tie,kim2021st}. The SLU network then has a cross-modal shared embedding space derived from both representations of ASR and NLU networks.

\vspace{-4mm}
\subsection{Network Integration with Interface}
These approaches were proposed to combine ASR and NLU networks and train the models in an end-to-end manner. 
They extracted acoustic information from the ASR network, including logits, hidden features, or text, and feed them into the NLU network via Interface~\cite{saxon2021end}. 
If the Interface is not differentiable, additional modules such as Gumbel-softmax or attention gate are needed~\cite{rao2021mean,saxon2021end,chen2021top}.

\vspace{-2mm}
\section{Proposed Method}

\begin{figure*}[ht!]
\centering
\includegraphics[height=7cm]{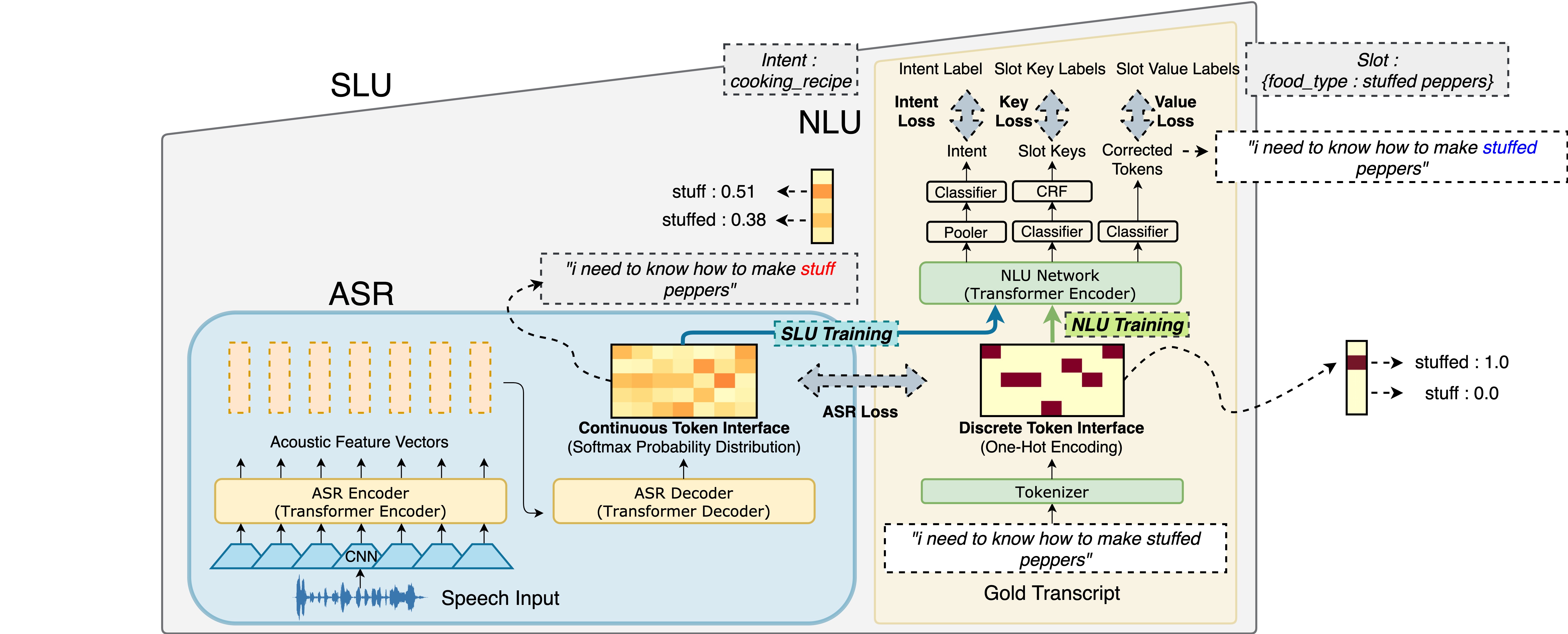}
\caption{Our proposed model architecture. We integrate the ASR network and the NLU network with \interface. 
We can train the entire network in an end-to-end manner or each component independently.}
\label{fig:propose_network_detail} 
\vspace{-5mm}
\end{figure*}

\vspace{-2mm}
\subsection{Integration Method and Model Components}
The SLU network we proposed consists of two main components: ASR and NLU networks.
For simple and robust integration via CTI, both networks are pre-trained with the same vocabulary. 
Given speech input $S$, the ASR network produces tokens with a continuous probability distribution by $softmax$ function and delivers the output vectors, $Z$ to the NLU network directly as in Eq.~(\ref{eq1}).
The NLU network finally outputs $Y$(intent, slot key, and slot value) as in Eq.~(\ref{eq2}).

\vspace{-3mm}
\begin{equation}
  Z = softmax(ASR_{Dec}(ASR_{Enc}(S)))
  \label{eq1}   
\end{equation}
\vspace{-4mm}
\begin{equation}
  Y = NLU(Z)
  \label{eq2}
\end{equation}
\vspace{-4mm}

We call this Z (the output of the ASR network) as CTI, which are considered few-hot encoded vectors from the point of view of the NLU network.
We hypothesize that it is noisy but more informative than one-hot encoded vectors generated by the conventionally used $argmax$ function.

If there is a confusing case between two tokens, the conventional ASR network makes only true/false binary predictions. 
In our proposed model, however, both pieces of information can be processed at the same time even with some noise.
We may think of this as a richer feature that contains information about how the token sounds, rather than just noise.

As a result, the entire SLU network can be trained in an end-to-end manner, with the intermediate noisy text representation.
Fig. \ref{fig:propose_network_detail} shows our proposed model.

\vspace{-4mm}
\subsubsection{ASR Network : \asr{}}
Our ASR network is a Sequence-to-Sequence (Seq2Seq) model consisting of Wav2Vec 2.0 Encoder~\cite{baevski2020wav2vec} and Transformer Decoder~\cite{vaswani2017attention}, where the Wav2Vec 2.0 Encoder is known to have remarkable representation power by contrastive learning task. 

\vspace{-4mm}
\subsubsection{NLU Network : \nlu}
We use \nlu{}~\cite{liu2019roberta}, one of the popular PLMs as our NLU network. 
We added an attention pooling layer, softmax classifiers, and a Conditional Random Field (CRF) at the end of the NLU network to predict intent, slot key, and slot value each.
We assume that our NLU network, trained with MLM, is suitable for understanding the noisy output of the ASR network.
Because this NLU network has the ability to reconstruct randomly masked noisy sentences, it can utilize noisy text representations without additional adaptation steps.

\vspace{-4mm}
\subsection{Multi-Task Learning Losses}

First, our proposed SLU network can be trained in an end-to-end manner with SLU loss ($L_{SLU}$) from speech-to-label data because it is directly integrated with the CTI.
With this Interface, we can train each component of the SLU network independently, even after integration. 
Thus, the ASR network can be trained with usual ASR loss ($L_{ASR}$) on speech-to-text data, and the NLU network can be trained with NLU loss ($L_{NLU}$) on text-to-label data.
And we can use all the losses to train the entire SLU network.

\vspace{-4mm}
\begin{equation}
  L_{Total} = L_{SLU} + L_{NLU} + L_{ASR}
  \label{eq_loss_total}   
\end{equation}
\vspace{-5mm}

\vspace{-4mm}
\subsubsection{SLU Loss}

The SLU loss ($L_{SLU}$) consists of Speech-to-Intent Loss ($L_{S2I}$), Speech-to-Slot Key Loss ($L_{S2K}$) and Speech-to-Slot Value Loss ($L_{S2V}$).
$L_{S2I}$ and $L_{S2K}$ are all Cross Entropy (CE) losses for intent, slot key classification each.
For $L_{S2K}$, we adopt in/out/begin (IOB) formats and pre-process the data to solve the misalignment between token and labels by the tokenizer~\cite{mesnil2014using}.
And $L_{S2V}$ loss is a summation of CE loss for each token.


\vspace{-4mm}
\subsubsection{NLU Loss}
The NLU loss ($L_{NLU}$) consists of Text-to-Intent Loss ($L_{T2I}$), Text-to-Slot Key Loss ($L_{T2K}$) and Text-to-Slot Value Loss ($L_{T2V}$).
The NLU loss is similar to the SLU loss, but the input modality is text.
Like the MLM task, we randomly mask some tokens of the gold transcript and put this masked sentence as input since it is too easy for the NLU network to predict the slot value from the pure sentence.


\vspace{-4mm}
\subsubsection{ASR Loss}
The ASR loss ($L_{ASR}$) is a token classification loss that is commonly used for Seq2Seq ASR network~\cite{chan2016listen}. 
Without this, the quality of the ASR hypothesis slowly degrades in the process of E2E SLU training.

\vspace{-2mm}
\section{Experiments}

\vspace{-4mm}
\subsection{Dataset}

Even though there are some SLU benchmark datasets, such as Fluent Speech Commands (FSC)~\cite{Lugosch2019}, their semantic complexity is insufficient to evaluate the capability of the NLU network~\cite{saxon2021end,mckenna2020semantic}.
The E2E SLU baseline without the LM, Wav2Vec2.0-Classifier, outperforms the other models with state-of-the-art performance, as shown in Table. \ref{tab:FSC}.
Therefore, we evaluate our proposed model on a more challenging dataset, SLURP~\cite{bastianelli2020slurp}, recently proposed for developing an in-home personal robot assistant. 

\vspace{-4mm}
\subsection{Training}


Firstly, we pre-train our ASR network on the LibriSpeech dataset~\cite{panayotov2015librispeech}. 
Here we use the pre-trained parameters of Wav2Vec 2.0 Encoder and Transformer Decoder with randomly initialized parameters.
As with the SLURP paper, we fine-tune the ASR network on the SLURP dataset because it requires domain adaptation to far-range speech data.
The Word Error Rate (WER) of our ASR network is $16.67$, which is slightly lower than the $16.20$ of the ASR network in the SLURP paper.

We use pre-trained RoBERTa as the NLU network, and finally, we connect two components via our proposed method (CTI), then fine-tune the model on the SLURP dataset with multi-task losses.
Note that, unlike other E2E SLU methods, each part of our model can behave like a stand-alone model.
Therefore, we can train the NLU network with text-to-label data and the ASR network with speech-to-text data at will.
We leave 46 audio files out of 50,628 training audio files during the fine-tuning because there are mismatches between speech and labels.

\begin{table}[pt!] 
\begin{center} 
\caption{\label{tab:FSC}The comparison of the IC accuracy between our baseline and other models on the FSC dataset.}
\vspace{-2mm}
\begin{adjustbox}{width=0.8\linewidth}
    \begin{threeparttable}
    \begin{tabular}{l|cccc}
    \toprule
    
    ~~\(\text{Model (E2E SLU)}\)~~ & ~~\(\text{Input}\)~~ & ~~\(\text{Dev}\)~~ & ~~\(\text{Test}\)~~ \\
    \midrule
    ~~\(\text{Lugosh et al.~\cite{Lugosch2019}}\)~~ & ~~\(\text{Speech}\)~~ & - & $98.8$  \\
    ~~\(\text{Kim et al.~\cite{kim2021two}}\)~~ & ~~\(\text{Speech}\)~~ & $97.8$ & $99.7$  \\
    ~~\(\text{Qian et al.~\cite{qian2021speech}}\)~~ & ~~\(\text{Speech}\)~~ & - & 99.7 \\
    \midrule
    ~~\(\text{Wav2Vec2.0-Classifier (Ours)}\)~~ & ~~\(\text{Speech}\)~~ & $\mathbf{98.9}$ & $\mathbf{99.7}$  \\
    
    \bottomrule
    \end{tabular}
    \end{threeparttable}
\end{adjustbox}
\end{center}
\vspace{-10mm}
\end{table}

\vspace{-4mm}
\subsection{Results and Analysis}
We evaluate the performance of SLU models with the metric of IC accuracy and SLU-F1 score, where SLU-F1 is a new metric for slot filling tasks proposed with the SLURP dataset~\cite{bastianelli2020slurp}.
The scores are shown in Table. \ref{tab:SLURP}.
There are three types of models: NLU, SLU (Inference Only), and SLU (E2E Train).
The score of our NLU network is $87.73$, which is higher than $84.84$ of the NLU network from the SLURP paper in IC.
To experiment with SLU (Inference Only) case, these stand-alone NLU networks need to be fine-tuned on the SLURP dataset.
On the other hand, in SLU (E2E Train), the NLU networks are not fine-tuned independently on the SLURP dataset but fine-tuned in an end-to-end manner.
Wav2Vec2.0-Classifier only shows $76.6$ IC accuracy because it does not explicitly understand the language information.
Among all SLU experiments, our model with all multi-task losses achieves state-of-the-art scores, $82.93$ (IC) and $71.12$ (SLU-F1).

\vspace{-4mm}
\subsubsection{Effectiveness of CTI}
We hypothesize that CTI makes NLU networks take richer information about the words that the ASR network confuses; the confusing words are represented as an interpolation of embedding vectors in the NLU network.
We compare SLU models with DTI, Gumbel-Interface~\cite{rao2021mean}, and CTI.
For SLU (Inference Only), CTI shows meaningful improvement compared to DTI by 0.8 (IC), and for SLU (E2E Train), CTI outperforms Gumbel-Interface by 0.83 (IC) and 0.57 (SLU-F1) under the same conditions except for the Interface.
We assume that the comparison between Gumbel-Interface and CTI could be different from ~\cite{saxon2021end} according to the domain of dataset.
Moreover, we introduce more multi-task losses, such as $L_{NLU}$ compared to~\cite{saxon2021end} for training with each interface.

\begin{table}[pt!]
\begin{center} 
\caption{\label{tab:SLURP}The comparison of IC accuracy and SLU-F1 score between proposed models and baselines. ASR$\rightarrow$NLU means inference with DTI and ASR$\Rightarrow$NLU means inference with CTI. 
In SLU (E2E Train), A, S, and N means ASR loss, SLU loss, and NLU loss for each.}
\vspace{-2mm}
\begin{adjustbox}{width=1\linewidth}
    \begin{threeparttable}
    \begin{tabular}{l|cccc}
    
    \toprule
    
    ~~\(\text{Model Type}\)~~ & Model & ~~\(\text{Intent}\)~~ & ~~\(\text{SLU-F1}\)~~ \\
    \midrule
    \multirowcell{2}{NLU} & NLU~\cite{bastianelli2020slurp} &  $84.84$  & -   \\   
     & NLU (Ours) &  $87.73$  & $84.34$   \\  
    \midrule
    \multirowcell{3}{SLU\\(Inference Only)} & ASR$\rightarrow$NLU~\cite{bastianelli2020slurp}  & $78.33$   &   $70.84$ \\
     & ASR$\rightarrow$NLU (Ours)  & $80.37$   &  $70.23$ \\
     & ASR$\Rightarrow$NLU (Ours)   & $81.17$   &  $70.20$ \\
    \midrule
    \multirowcell{6}{SLU\\(E2E Train)} & Wav2Vec2.0-Classifier (Ours) & $76.6$  & -  \\
     & Gumbel-Interface (A+S+N)~\cite{rao2021mean} & 82.10 & 70.55   \\
     & CTI (A+S) & $82.39$  & $70.61$  \\
     & CTI (A+S+N) & $\mathbf{82.93}$  & $\mathbf{71.12}$   \\
    \cmidrule{2-4}
     & CTI (A+S+N) + Extra data (Text-only) & $84.34$  & $71.08$   \\
     & CTI (A+S+N) + Extra data (All)  & $\mathbf{86.92}$  & $\mathbf{74.66}$   \\
    \bottomrule
    \end{tabular}
    \end{threeparttable}
    
\end{adjustbox}
\end{center}
\vspace{-10mm}
\end{table}

\vspace{-4mm}
\subsubsection{NLU Loss}
We train our model with and without $L_{NLU}$ because CTI allows NLU networks to be trained independently on text-only data.
Adding $L_{NLU}$ increases the model score by $0.54$ (IC) and $0.51$ (SLU-F1).
This shows that CTI can preserve the NLU network's powerful text representation, which is learned in the pre-training procedure.

\vspace{-4mm}
\subsubsection{Extra Data : SLURP-Synth}
We conduct an additional experiment to check if using extra data improves the model performance as studied in~\cite{lugosch2020using}.
The extra data is SLURP-Synth where it consist of text, labels and synthesized speech by Google’s Text-to-Speech system~\cite{bastianelli2020slurp}. 
Here we leave 24,365 files that have out-of-distribution labels among the 69,253 files.

We first train the entire network with additional Text-only training data. To do this, we extract the Transcript and Intent pairs from SLURP-synth.
The results show that using additional Text-only data increases the IC by 2.5.
Second, we train the entire network with SLURP-Synth data containing all text, labels, and synthesized speech.
This increase the score by $2.58$ (IC) and $3.58$ (SLU-F1) than Extra data (Text-only) case, and we finally achieved $86.92$ (IC) and $74.66$ (SLU-F1) which is the state-of-the-art score.
This means that critical information still remains in speech-to-label training that has not been fully discovered in text-to-label training.

\vspace{-2mm}
\section{Conclusions and Future Work}
We propose a simple and robust method for integrating two pre-trained networks via novel Interface, CTI. 
We achieve state-of-the-art scores of 82.93 (IC), 71.12 (SLU-F1) on the SLURP dataset, and 86.92 (IC), 74.66 (SLU-F1) when we add the SLURP-Synth dataset.
In the future, we plan to investigate a new pre-training strategy in which the NLU recovers some tokens corrupted by acoustic noise, such as phoneme-level ASR error distributions.
Then the noise pattern of the NLU network inputs will match the ASR network outputs; the E2E SLU model will be less susceptible to the problem of representation gap.

\clearpage


\let\OLDthebibliography\thebibliography
\renewcommand\thebibliography[1]{
  \OLDthebibliography{#1}
  \setlength{\parskip}{0pt}
  \setlength{\itemsep}{0pt plus 0.4ex}
}

\bibliographystyle{IEEEbib}
\bibliography{220216_icassp_final_integration_seo_kwak_lee}

\end{document}